\documentclass[lettersize,journal]{IEEEtran}
\usepackage{amsmath,amsfonts}
\usepackage{algorithmic}
\usepackage{array}
\usepackage[caption=false,font=normalsize,labelfont=sf,textfont=sf]{subfig}
\usepackage{textcomp}
\usepackage{stfloats}
\usepackage{url}
\usepackage{verbatim}
\usepackage{graphicx}
\usepackage{xcolor}
\graphicspath{{figs/}}
\def\BibTeX{{\rm B\kern-.05em{\sc i\kern-.025em b}\kern-.08em
    T\kern-.1667em\lower.7ex\hbox{E}\kern-.125emX}}
\usepackage{balance}
\begin{document}
\title{SDI-Net: Toward Sufficient Dual-View Interaction for Low-light Stereo Image Enhancement}
\author{\IEEEauthorblockN{LinLin Hu\IEEEauthorrefmark{2}, Ao Sun\IEEEauthorrefmark{2}, Shijie Hao\IEEEauthorrefmark{1}\IEEEauthorrefmark{2}}, Richang Hong\IEEEauthorrefmark{2}, Meng Wang\IEEEauthorrefmark{2}\\
    \IEEEauthorblockA{\IEEEauthorrefmark{2}Hefei University of Technology\\ }
}


\maketitle
\begin{figure*}[h]
	\centering
	\includegraphics[scale=0.55]{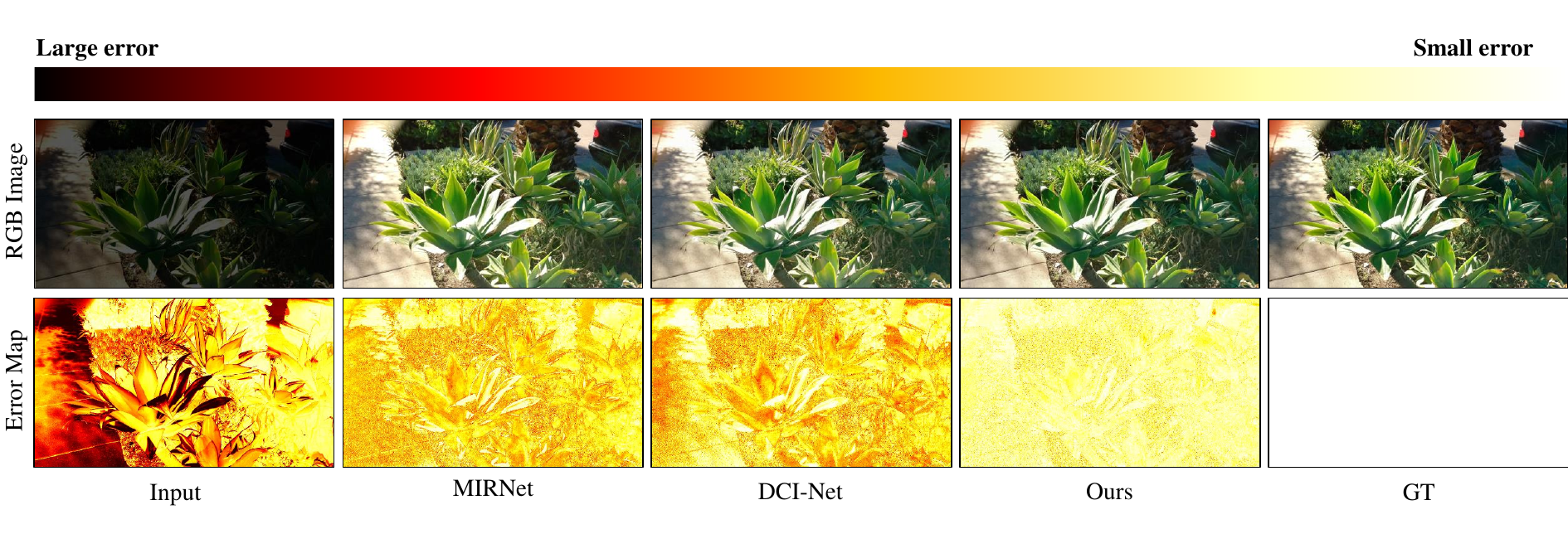}
	\vspace{-6mm}
	\caption{An example of low-light image enhancement based on the recently proposed monocular (single image) low-light enhancement method MIRNet\cite{zamir2020learning} and binocular (stereo image) low-light enhancement method DCI-Net\cite{zheng2023decoupled}, and our SDI-Net. The error maps represent the performance on restoring image details, in which darker pixels indicate larger errors from the ground-truth (GT) image. The image is from the Holopix50k dataset.}
	\label{fig1}
	\vspace{-3mm}
\end{figure*}
\begin{abstract}
Currently, most low-light image enhancement methods only consider information from a single view, neglecting the correlation between cross-view information. Therefore, the enhancement results produced by these methods are often unsatisfactory. In this context, there have been efforts to develop methods specifically for low-light stereo image enhancement. These methods take into account the cross-view disparities and enable interaction between the left and right views, leading to improved performance. However, these methods still do not fully exploit the interaction between left and right view information. To address this issue, we propose a model called Toward Sufficient Dual-View Interaction for Low-light Stereo Image Enhancement (SDI-Net). The backbone structure of SDI-Net is two encoder-decoder pairs, which are used to learn the mapping function from low-light images to normal-light images. Among the encoders and the decoders, we design a module named Cross-View Sufficient Interaction Module (CSIM), aiming to fully exploit the correlations between the binocular views via the attention mechanism. The quantitative and visual results on public datasets validate the superiority of our method over other related methods. Ablation studies also demonstrate the effectiveness of the key elements in our model.

\end{abstract}


\section{Introduction}
Low-light image enhancement aims to obtain images with good visibility from low-light images, which has been a main low-level vision task in both academic and industrial communities. The research on low-light image enhancement has made great progress in the past decades. Conventional non-learning-based methods \cite{guo2016lime, coltuc2006exact, ibrahim2007brightness, lee2013contrast, land1977retinex, fu2016weighted, li2018structure, yuan2012automatic, yue2017contrast} for low-light enhancement primarily concentrate on brightness adjustment and contrast enhancement based on prior knowledge such as the Retinex theory. However, these approaches frequently yield suboptimal enhancement results, often resulting in less satisfying visual quality. In comparison to the conventional methods, low-light image enhancement methods based on deep neural network \cite{jiang2021enlightengan,yang2020fidelity,liu2021retinex,ma2022toward,zhang2019kindling,Chen2018Retinex} have achieved significant advancements in recent years. These methods employ convolutional neural networks (CNNs) or generative adversarial networks (GANs) as the backbone framework to learn the mapping function from low-light images to normal light images.  

The above-mentioned models are oriented for monocular images. In another word, the input of these models is one single image. In recent years, the emergence of stereo cameras has sparked interest in stereo vision across diverse fields, which provides richer information than monocular image processing systems. For example, as for binocular images, there exist sufficient horizontal discrepancies between the left view and the right view of a same object. Based on this characteristic, many works have been dedicated to developing stereo image restoration techniques for enhancing image visual quality\cite{chu2022nafssr, wang2019learning, wang2021symmetric}, such as stereo super-resolution, stereo deblurring, and stereo dehazing. The primary challenge faced by stereo image restoration methods is how to fully leverage the correlations between the two views, specifically, how to effectively associate the left and the right views for obtaining more effective feature representation. For example, for the stereo image super-resolution task, iPASSENet\cite{wang2021symmetric} aims to capture the correspondence between the left and the right views through modeling the parallax attention mechanism. 

As a typical image restoration task, low-light image enhancement is also striving for advancements by incorporating stereo vision\cite{huang2022low, jung2020multi, liao2022no}. In comparison to the monocular (single-image) low-light enhancement methods, low-light stereo image enhancement (LLSIE) methods obtain better performance in general. However, due to the problem that low-light images suffer from low contrast and imaging noise, the current LLSIE research is still limited in well restoring image details. For example, as shown in Fig.\ref{fig1}, the result of LLSIE method DCI-Net\cite{zheng2023decoupled} also has restoring errors as large as the single-image low-light enhancement method MIRNet\cite{zamir2020learning}. One of the main reasons is that the interaction between the left and right views of the current LLSIE methods is not sufficient.

To address this problem, we propose SDI-Net aiming at fully exploring dual-view interaction for low-light stereo image enhancement. The proposed model is able to restore the illumination and details of low-light images through sufficient interaction between the left and right views. We develop an intermediate interaction module named Cross-View Sufficient Interaction Module (CSIM) to explore and strengthen the interaction between features of both views. The first part of CSIM is the Cross-View Attention Interaction Module (CAIM), which is adept at calculating the disparity between the left and right views and aligning them with high precision. The second part comprises a Pixel and Channel Attention Block (PCAB), designed to differentially enhance areas of varying brightness levels, thereby restoring richer details and textures. The proposed SDI-Net is highlighted in the following aspects:
\begin{itemize}
\item SDI-Net adopts two identical UNets as its backbone structure. Each UNet acts as the image encoder and decoder for either of the left-view image and the right-view image. This symmetry model structure ensures the two encoder-decoder branches have the same learned feature representations at different levels, facilitating sufficient interactions between the learning pipelines of the two views. 

\item Between the encoders and decoders of SDI-Net, we design the Cross-View Sufficient Interaction Module (CSIM) to make the learned feature representations fully interact with each other at different aspects.

\item SDI-Net has superior performance on the Middlebury dataset and the Holopix50k dataset over other low-light stereo image enhancement methods, as well as several representative single-image low-light image enhancement methods.

\end{itemize}

The rest of this paper is organized as follows. Section II briefly introduces the related works. In Section III, we describe the overview and details of the proposed SDI-Net. Section IV reports the experimental results on two public datasets. Section V finally concludes the paper.


\section{Related Works}

\subsection{Single-image low-light enhancement methods}

The traditional single-image low-light enhancement methods mainly include the histogram equalization and retinex-based methods. Histogram equalization methods directly adjust the dynamic range of low-light images, thereby enhancing image contrast\cite{lee2013contrast}. Retinex-based methods aim to decompose a low-light image into reflection and illumination layers, and adjust the brightness of the illumination layer to achieve the low-light enhancement\cite{cai2017joint, li2018structure, ren2020lr3m}. These methods are highlighted for their clear physical interpretability. However, they are limited in its capability of fitting the complex mapping function between low-light and normal-light images. By leveraging deep learning models, learning-based low-light enhancement methods well solve this limitation and greatly improve the performance. These methods are able learn end-to-end appearance mapping functions between low-light inputs and normal-light outputs\cite{Chen2018Retinex,lim2020dslr, ren2019low, zamir2020learning}. Recently, models with limited or no supervision also emerge as popular low-light enhancement methods \cite{guo2020zero, yang2020fidelity,li2021learning, ma2022toward} due to their lightweight and fast characteristics. However, they are prone to introduce over-enhancement and obvious artifacts into enhancing results. Despite of the success of single-image low-light enhancement methods, their information source is one single input all along. In contrast, methods taking stereo images as inputs pave way for utilizing richer information in this task, which have the potential of achieving better enhancing performance.

\subsection{Stereo image restoration methods}
Recently, stereo image restoration methods have begun to attract more attention, and obtained better performance than single-image restoration methods, such as the tasks of super-resolution, deblurring, deraining, dehazing, and low-light enhancement. In the stereo image super-resolution method, Jeoh et al.\cite{jeon2018enhancing} propose the first stereo super-resolution network to compensate for the parallax between stereo images by shifting. Wang et al.\cite{wang2019learning} use a parallax attention mechanism to merge similar features in two views to explore pixel correspondence. In the stereo deblurring method, Zhou et al.\cite{zhou2019davanet} align features by estimating the difference between the left and right views. Li et al.\cite{li2022learning} propose a new stereo image deblurring model by exploring two-pixel alignment. In the stereo dehazing method, Pang et al.\cite{pang2020bidnet} use a stereo transformation module to explore the correlation between binocular images. As for our low-light task, DVENet\cite{huang2022low} is regarded as the representative method, which is to enhance the network by integrating multi-scale dual-view features, and fuse the features of a single image under the guidance of the light map. Recently, DCI-Net\cite{zheng2023decoupled} tries to further exploits the interaction between the left and right views by considering the spatial connection between multiple scales. 

In this paper, we focus on the low-light enhancement task. We propose SDI-Net to fully incorporate the information from the left and right views via the attention mechanism imposed on different aspects, such as the view level, the channel level and the pixel level. We introduce the technical details of SDI-Net in the following section.

\begin{figure*}[t]
	\centering
	\includegraphics[scale=0.46]{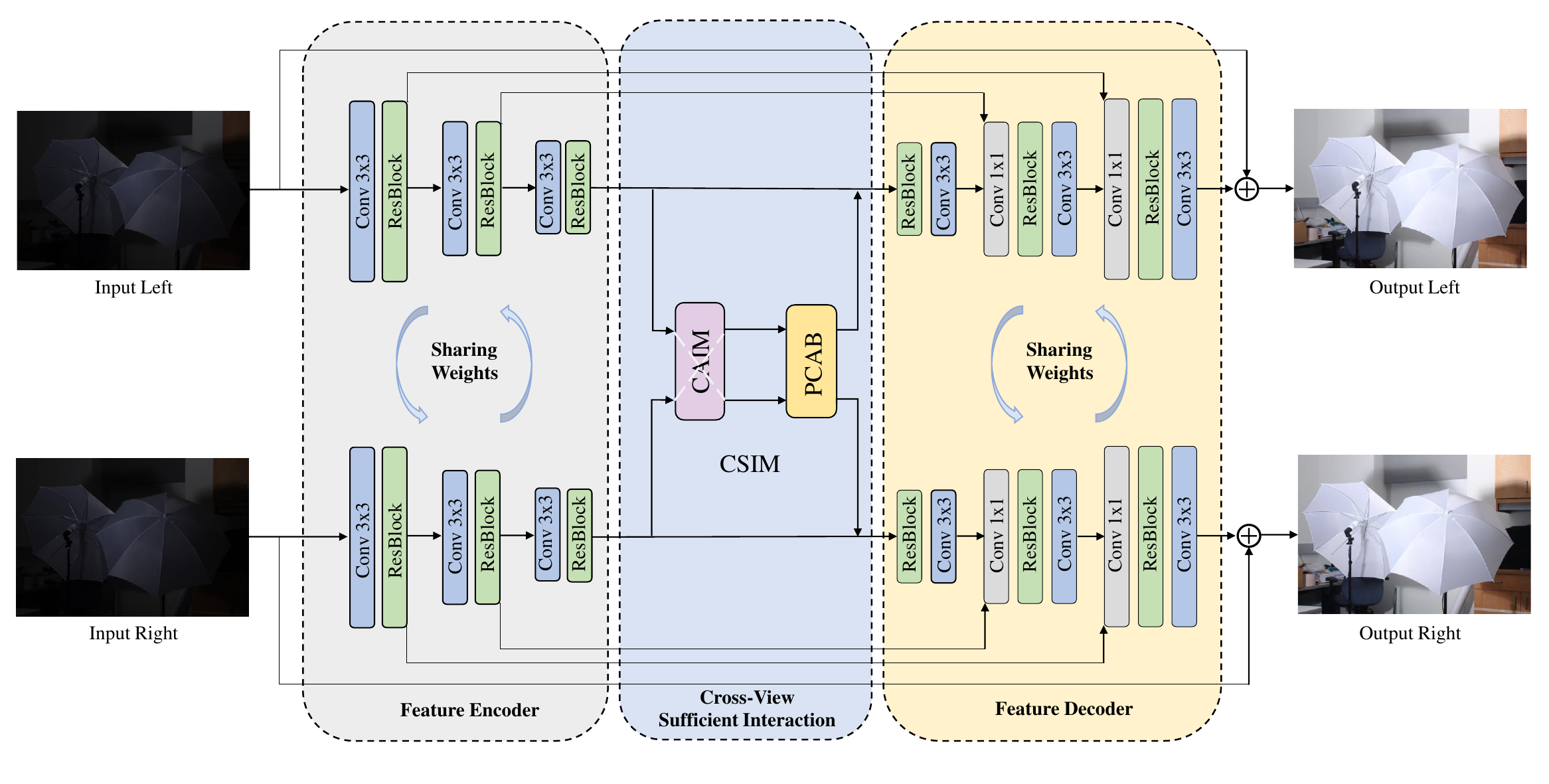}
	\vspace{-8mm}
	\caption{The overall framework of the proposed SDI-Net, which is consisted of three stages, i.e., Feature Encoder, Cross-View Sufficient Interaction, and Feature Decoder.}
	\label{fig2}
	\vspace{-4mm}
\end{figure*}

\section{Method}
The SDI-Net model takes low-light stereo image pairs as inputs, and utilizes two identical UNet branches as the backbone structure to learn the mapping from low-light stereo images to normal-light stereo images. Between the encoders and the decoders of the two branches, we introduce Cross-View Sufficient Interaction Module (CSIM) to refine the learned image feature representation. In the following, we describe the overall framework, the CSIM module, and the employed loss functions.

\begin{figure*}[t]
	\centering
	\includegraphics[scale=0.47]{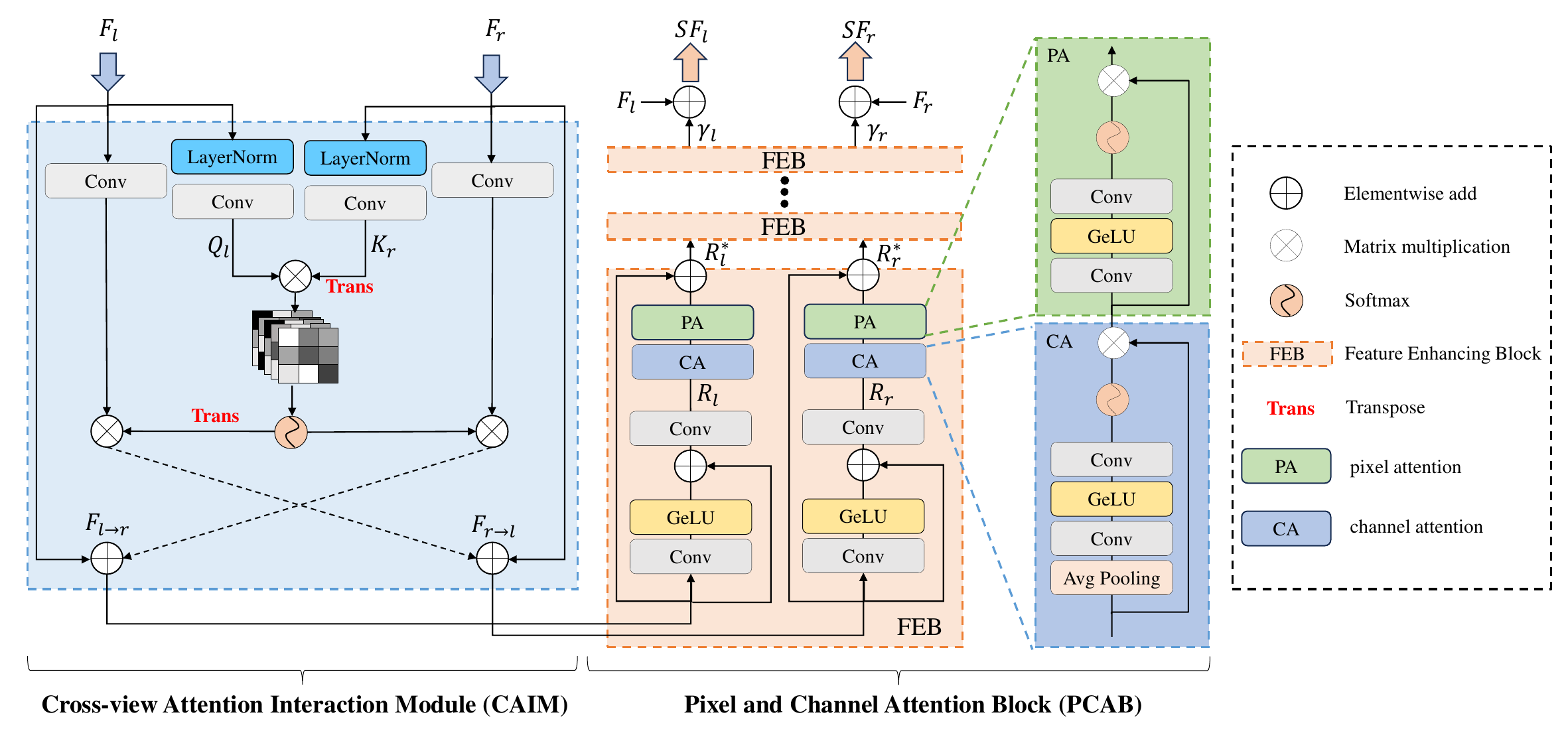}
	\vspace{-8mm}
	\caption{The architecture of Cross-View Sufficient Interaction Module (CSIM). It contains Cross-view Attention Interaction Modul (CAIM) and Pixel and Channel Attention Block (PCAB)}
	\label{fig3}
	\vspace{-3mm}
\end{figure*}

\subsection{Overall framework}\label{AA}
The overall framework of SDI-Net is shown in Fig.\ref{fig2}, which can be divided into three stages: \textbf{Feature Encoder}, \textbf{Cross-View Sufficient
Interaction}, and \textbf{Feature Decoder}.

\textbf{Feature Encoder.} First, we take the left-view low-light image ${I_l} \in {\mathbb{R}^{H \times W \times 3}}$ and the right-view low-light image ${I_r} \in {\mathbb{R}^{H \times W \times 3}}$ as the inputs of the encoders. Based on the encoders, the feature maps ${F_l} \in {\mathbb{R}^{\frac{H}{4} \times \frac{W}{4} \times 3}}$ and ${F_r} \in {\mathbb{R}^{\frac{H}{4} \times \frac{W}{4} \times 3}}$ are learned. This process can be formulated as:
\begin{equation}
{F_l} = FE({I_l}),{F_r} = FE({I_r})\label{eq}
\end{equation}
where $FE( \cdot )$ means feature encoder.  

The encoders comprise several convolutional layers followed by down-sampling operations, aiming to initially capture and encode the local and global information of each view. Of note, the learned network weights are shared between the two branches. Then, we send the obtained feature information to the next stage of cross-view sufficient interaction.

\textbf{Cross-View Sufficient Interaction Module.} The CSIM module stays between the encoders and the decoders of the two UNet branches, aiming to promote sufficient interaction between the features learned from the two views. The CSIM module is composed of the Cross-View Attention Interaction Module (CAIM) part and the Pixel and Channel Attention Block (PCAB) part. The former part focuses on exploring the mutual attention of ${F_l}$ and ${F_r}$ at the view level. Then, the later part concentrates on exploiting the attention mechanism at the channel level and the pixel level. Specifically, channel attention ($CA$) helps to emphasize informative channels, while pixel attention ($PA$) highlights relevant spatial locations. In this way, the CSIM module enables sufficient interaction between the two views, obtaining the mutually fused features $S{F_l} \in {\mathbb{R}^{\frac{H}{4} \times \frac{W}{4} \times 3}}$ and $S{F_r} \in {\mathbb{R}^{\frac{H}{4} \times \frac{W}{4} \times 3}}$:

\begin{equation}
S{F_l},S{F_r} = CSIM({F_l},{F_r})\label{eq}
\end{equation}
where $CSIM( \cdot )$ represents the core part of our SDI-Net. Then, $S{F_l}$ and $S{F_r}$ are sent to the next feature decoding stage for image reconstruction.

\textbf{Feature Decoder.} In this stage, the feature decoders reconstruct the enhanced stereo images based on the enriched feature representation $S{F_l}$ and $S{F_r}$. The two feature maps are respectively fed into the decoders of the two UNet branches. This process involves two up-sampling operations, followed by convolution layers to recover the spatial dimensions and generate the final enhanced stereo images. The first up-sampling operation is to send $S{F_l} \in {\mathbb{R}^{\frac{H}{4} \times \frac{W}{4} \times 3}}$ and $S{F_r} \in {\mathbb{R}^{\frac{H}{4} \times \frac{W}{4} \times 3}}$ into the convolutional layer through 8 residual blocks and one 3*3 up-sampling convolutional layer, respectively, to obtain the first recovery of the middle layer features ${M_l} \in {\mathbb{R}^{\frac{H}{2} \times \frac{W}{2} \times 3}}$ and ${M_r} \in {\mathbb{R}^{\frac{H}{2} \times \frac{W}{2} \times 3}}$, and then fuse the feature information extracted by the first layer of the Feature Encoder on the spatial domain through the concatenation operation. The second up-sampling operation is to process ${M_l} \in {\mathbb{R}^{\frac{H}{2} \times \frac{W}{2} \times 3}}$ and ${M_r} \in {\mathbb{R}^{\frac{H}{2} \times \frac{W}{2} \times 3}}$ identically to obtain ${E_l} \in {\mathbb{R}^{H \times W \times 3}}$ and ${E_r} \in {\mathbb{R}^{H \times W \times 3}}$. This process can be formulated as follows:
\begin{equation}
{E_l} = FD(S{F_l}),{E_r} = FD(S{F_r})\label{eq}
\end{equation}
where $FD( \cdot )$ stands for the feature decoders. Similar as the encoding stage, the learned weights of the decoders are shared between the two branches. 

To train the whole model, we use the L1 loss and the FFT loss to achieve the low-light enhancement task.

\subsection{Cross-View Sufficient Interaction Module (CSIM)}
The CSIM module is composed of two parts, i.e. Cross-View Attention Interaction Module (the left part of Fig.\ref{fig3}) and Pixel and Channel Attention Block (the right part of Fig.\ref{fig3}), which are described in the following.

\textbf{Cross-View Attention Interaction Module (CAIM)}. CAIM is used to extract the correlation information at the view level. The correlation computing process is based on Scaled Dot Product Attention\cite{vaswani2017attention}, which utilizes all the keys to compute the dot products of the query, and applies the softmax function to obtain the weights of the values:
\begin{equation}
Attention(Q,K,V) = softmax (\frac{{Q{K^T}}}{{\sqrt C }})V\label{eq}
\end{equation}
where $Q$ represents the query matrix, $K$ represents the key matrix, $V$ represents the value matrix, and $C$ is the channel numbers. In our application, we use ${Q_l}=Conv(LN({F_l}))$ and ${K_r}=Conv(LN({F_r}))$ to represent the features of the two views, where $LN( \cdot )$ represents layer normalization, and $Conv( \cdot )$ represents a 1×1 convolution operation. The refined features can be obtained as:  
\begin{equation}
{F_{r \to l}} = Attention({Q_l},{K_r},Conv({F_l}))+F_l\label{eq}
\end{equation}
\begin{equation}
{F_{l \to r}} = Attention({K_r},{Q_l},Conv({F_r}))+F_r\label{eq}
\end{equation} 
From above process, the computing process for obtaining ${F_{r \to l}}$ and ${F_{l \to r}}$ fully considers the interaction from each other view.

\textbf{Pixel and Channel Attention Block (PCAB)}. PCAB is composed on several stacked feature enhancing blocks for further refining ${F_{r \to l}}$ and ${F_{l \to r}}$. In the first feature enhancing block (FEB), we initially pre-process ${F_{r \to l}}$ and ${F_{l \to r}}$ based on the following procedure: 
\begin{equation}
{R_l} = Conv(G(Conv({F_{r \to l}})) + {F_{r \to l}})\label{eq1}
\end{equation}
\begin{equation}
{R_r} = Conv(G(Conv({F_{l \to r}})) + {F_{l \to r}})\label{eq2}
\end{equation} 
where $Conv( \cdot )$ refers to a 3×3 convolution operation and $G( \cdot )$ stands for GeLU activation function. Then, the channel attention function $CA( \cdot )$ and the pixel attention function $PA( \cdot )$ \cite{qin2020ffa} are introduced to further exploit the feature correlation at the channel level and the pixel level:
\begin{equation}
{R_l}^*= PA(CA({R_l})),   {R_r}^* =PA(CA({R_r}))\label{eq3}
\end{equation} 
The channel attention ($CA( \cdot )$) learns the importance of each feature channel, aiming to highlight more useful information. The pixel attention ($PA( \cdot )$) dynamically adjust the importance of each pixel, which aiming to enhance important details and texture features while reducing the impact from noise.

In the following, the above process is repeated by stacking multiple FEBs in a sequential way. Of note, instead of ${F_{r \to l}}$ and ${F_{l \to r}}$, the inputs of the blocks other than the first FEB are the previous neighboring FEB's outputs. In our research, we empirically set the number of FEBs as 10 to build the PCAB module.

At the end of PCAB, we combine the gradually refined ${R_l}^*$ and ${R_r}^*$ with its original feature representation ${F_l}$ and ${F_r}$ using a weighted element-wise addition to obtain $S{F_l}$ and $S{F_r}$, which is the final feature representation learned by the CSIM module:
\begin{equation}
S{F_l} = {\gamma _l}{R_l}^* + {F_l}\label{eq}
\end{equation} 
\begin{equation}
S{F_r} = {\gamma _r}{R_r}^* + {F_r} \label{eq}
\end{equation} 
where $\gamma _l$ and $\gamma _r$ are trainable weights and are initialized with zeros to ensure stable training. 

\begin{table*}[!ht]
\caption{Quantitative performance on the Middlebury and Synthetic Holopix50k datasets. The evaluation metrics of PSNR and SSIM are used. The best results are highlighted in bold.}
    \centering
    \renewcommand{\arraystretch}{1.2}
    \resizebox{\linewidth}{!}{
    \tiny
    \begin{tabular}{ccccccc}
    \hline
        \multicolumn{2}{c}{Methods}& Venue & \multicolumn{2}{c}{Middlebury} & \multicolumn{2}{c}{Holopix50}  \\ \hline
        ~ & ~ & ~ & Left & Right & Left & Right \\ \hline
        ~ & NPE\cite{wang2013naturalness} & tip’13 & 15.05 / 0.6778 & 15.20 / 0.6790 & 19.06 / 0.7813 & 19.18 / 0.7806 \\ 
        ~ & LIME\cite{guo2016lime} & tip’17 & 14.19 / 0.8442 & 14.41 / 0.8424 & 18.85 / 0.8361 & 18.91 / 0.8359 \\ 
        ~ & JieP\cite{cai2017joint} & iccv’17 & 9.97 / 0.6476 & 10.08 / 0.6463 & 13.43 / 0.6943 & 13.43 / 0.6903 \\
        ~ & RRM\cite{li2018structure} & tip’18 & 11.20 / 0.7111 & 11.33 / 0.7084 & 15.04 / 0.7210 & 15.03 / 0.7191 \\ 
        ~ & ZeroDCE\cite{li2021learning} & tpami’20 & 15.66 / 0.7231 & 15.21 / 0.7063 & 12.26 / 0.6277 & 14.30 / 0.6767 \\ 
        ~ & RetinexNet\cite{Chen2018Retinex} & bmvc’18 & 22.62 / 0.8047 & 22.69 / 0.7955 & 19.19 / 0.7800 & 19.20 / 0.7788 \\ 
        Monocular & MBLLEN\cite{lv2018mbllen} & bmvc’18 & 18.87 / 0.8239 & 19.59 / 0.8201 & 20.32 / 0.8370 & 20.51 / 0.8403 \\ 
        ~ & DSLR\cite{lim2020dslr} & tmm’20 & 22.32 / 0.7430 & 22.37 / 0.8361 & 21.49 / 0.8125 & 21.62 / 0.8019 \\ 
        ~ & KIND\cite{zhang2019kindling} & acm mm’19 & 22.76 / 0.9297 & 21.39 / 0.9007 & 23.59 / 0.8956 & 24.27 / 0.8972 \\ 
        ~ & DRBN\cite{yang2020fidelity} & cvpr’20 & 30.80 / 0.9451 & 31.23 / 0.9408 & 24.82 / 0.9047 & 25.35 / 0.9017 \\
        ~ & SNR-Aware\cite{xu2022snr} & cvpr’22 & 32.84 / 0.9049  & 32.93 / 0.8942 & 24.51 / 0.7697 & 24.55 / 0.7684  \\
        ~ & LLFormer\cite{wang2023ultra} & aaai’23 & 31.85 / 0.9020  & 31.53 / 0.8891 & 23.36 / 0.7674 & 23.49 / 0.7557  \\
        ~ & MIRNet\cite{zamir2020learning} & tpami’23 & 31.86 / 0.9451 & 31.95 / 0.9415 & 25.39 / 0.9060 & 25.91 / 0.9102 \\ \hline
        ~ & iPASSRNet\cite{wang2021symmetric} & cvpr’21 & 34.07 / 0.9404 & 33.69 / 0.9336 & 26.28 / 0.8780 & 26.72 / 0.8770 \\ 
        Stereo & DVENet\cite{huang2022low} & tmm’22 & 35.47 / 0.9492 & 35.15 / 0.9441 & 27.70 / 0.9236 & 28.34 / \textbf{0.9265} \\ 
        ~& DCI-Net\cite{zheng2023decoupled} & acm mm’23 & 32.92 / 0.9297 & 32.30 / 0.8244 & 26.67 / 0.8930 & 27.07 / 0.9044 \\ 
        ~ & Ours & - & \textbf{35.91 / 0.9501} & \textbf{35.99 / 0.9458} & \textbf{28.72 / 0.9239} & \textbf{29.03 / 0.9265} \\ \hline
    \end{tabular}
    }
\label{table1}
\end{table*}

\subsection{Loss function}
In this paper, we use the L1 loss and the FFT loss to train the whole network, which can be expressed as follows:
\begin{equation}
L = {L_1} + \lambda {L_{fre}}\label{eq}
\end{equation}
where $\lambda$ stands for a hyper-parameter, empirically set to 0.1. It is important to note that ${L_{fre}}$ helps to restore the normal light image via preserving the frequency-domain image characteristics. The two loss terms can be expressed as:
\begin{equation}
  {L_1} = {\left\| {{E_l} - {E_l}^G} \right\|_1} + {\left\| {{E_r} - {E_r}^G} \right\|_1} \hfill  \label{eq}
\end{equation}
\begin{equation}
  {L_{fre}} = {\left\| {\varphi ({E_l}) - \varphi ({E_l}^G)} \right\|_1} + {\left\| {\varphi ({E_r}) - \varphi ({E_r}^G)} \right\|_1} \hfill \label{eq}
\end{equation}
 where ${\left\|  \cdot  \right\|_1}$ is the L1 loss, which is used to maintain the overall structure and normal-light and low-light details of the image. $\varphi ( \cdot )$ stands for Fast Fourier Transform, which helps to recover the texture details and edge information of the image, making the reconstructed image closer to the image under normal-light conditions. ${{E_l}}$ and ${{E_r}}$ represent the restored left and right normal-light images, and ${{E_l}^G}$ and ${{E_r}^G}$ represent the ground truth corresponding to ${{E_l}}$ and ${{E_r}}$.

\begin{figure*}[t]
	\centering
	\includegraphics[scale=0.55]{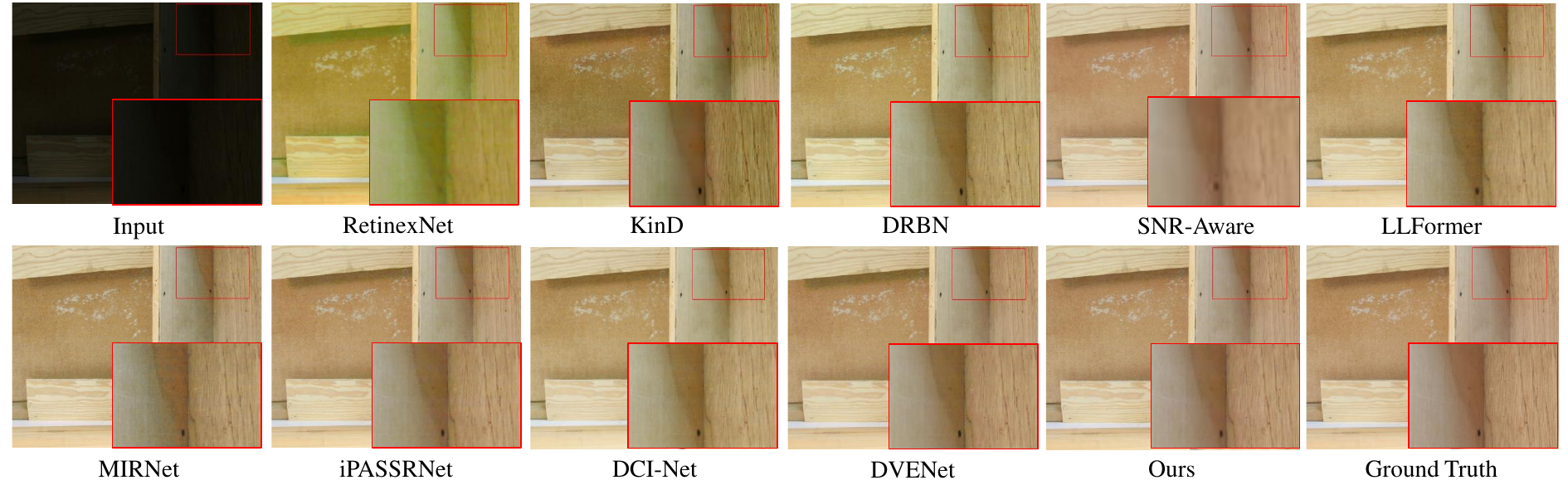}
	\vspace{-4mm}
	\caption{Visual comparison of the enhancement results on the Middlebury dataset. Better with a zoomed-in view.}
	\label{fig4}
	\vspace{-3mm}
\end{figure*}

\section{Experiments}
\subsection{Experiment settings}

\begin{figure*}[t]
	\centering
	\includegraphics[scale=0.55]{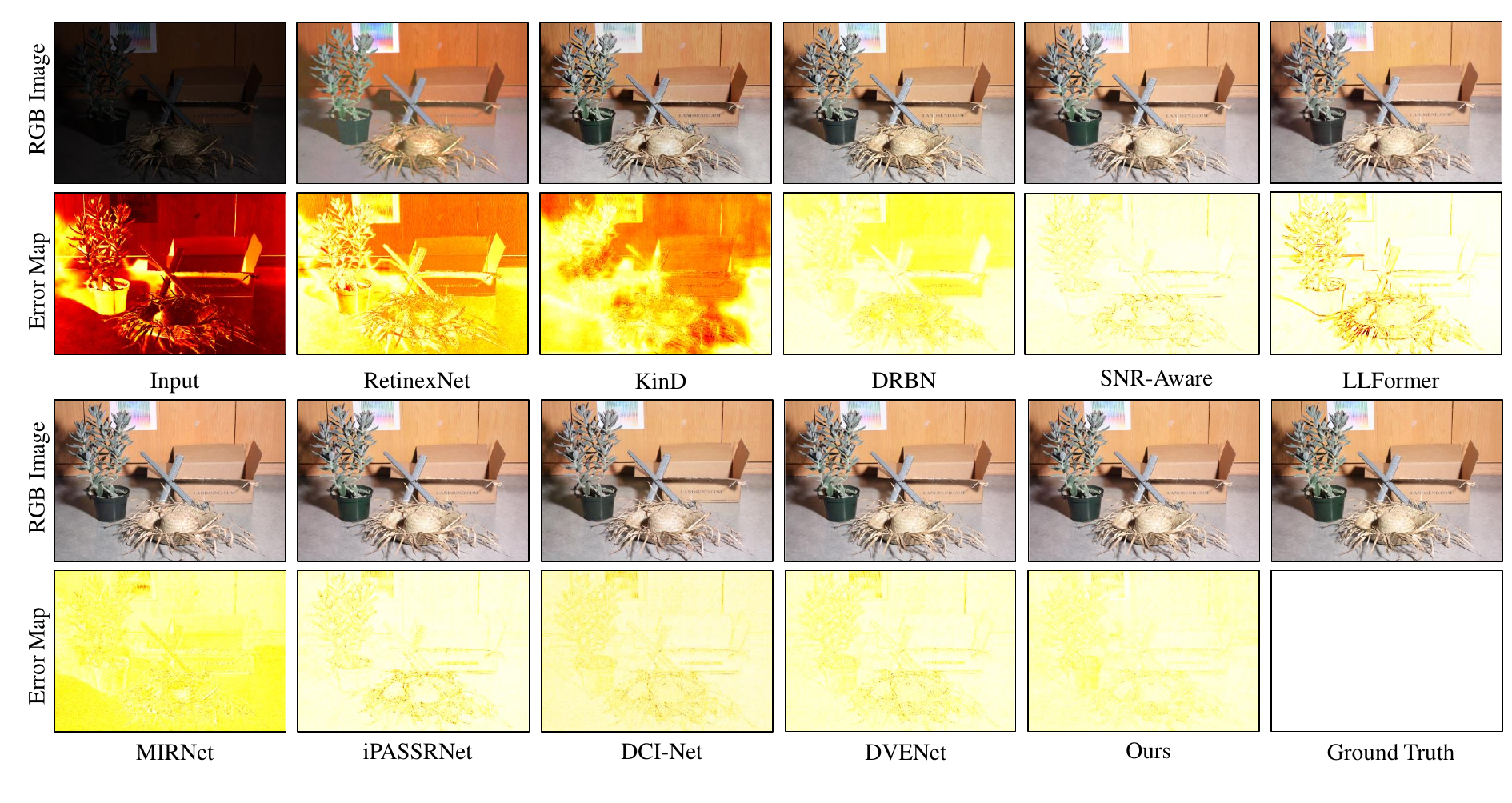}
	\vspace{-5mm}
	\caption{Visual comparison of the enhancement results on the Middlebury dataset. In the error maps, darker pixels indicate larger errors. Better with a zoomed-in view.}
	\label{fig5}
	\vspace{-3mm}
\end{figure*}
\begin{figure*}[t]
	\centering
	\includegraphics[scale=0.6]{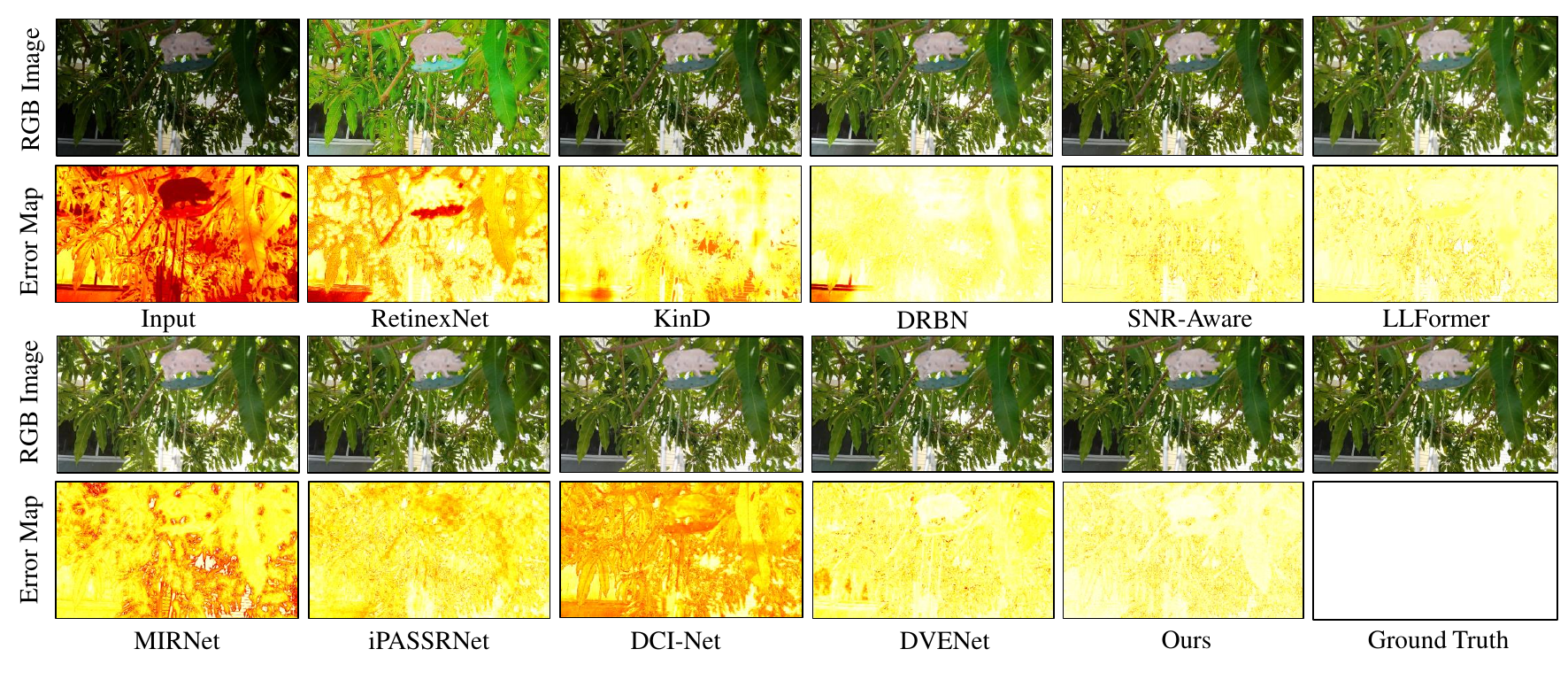}
	\vspace{-3mm}
	\caption{Visual comparison of the enhancement results on the Synthetic Holopix50k dataset. In the error maps, darker pixels indicate larger errors. Better with a zoomed-in view.}
	\label{fig6}
	\vspace{-2mm}
\end{figure*}

\textbf{Datasets}. We adopt the existing public datasets specifically designed for low-light stereo image enhancement, i.e., Middlebury and Synthetic Holopix50k from DVENet\cite{huang2022low}, to evaluate the effectiveness of our method. The Middlebury dataset selects 136 pairs of images as the training set with resolution of 512×512 and 36 pairs of images for testing. The Middlebury dataset is carefully built to offer high-quality, precisely calibrated stereo images, which are crucial for the low-light stereo image enhancement task. The Synthetic Holopix50k dataset is a much larger dataset. Its training set includes 1128 images with resolution of 512×512, and its test set includes 159 images. The Holopix50k dataset originates from a larger collection of real-world images captured by dual-camera smartphones, offering a broader but less controlled variety of low-light scenarios. This diversity is vital for training models to generalize across a wide range of real-world conditions.

\textbf{Evaluation Metrics}. We use two mostly-adopted full reference quality metrics PSNR and SSIM to evaluate the model performance. Higher PSNR or SSIM values indicate better performance.

\textbf{Methods for Comparison}. We compare our proposed SDI-Net with several single-image (monocular) low-light enhancement methods and stereo-image low-light enhancement methods. The single-image low-light enhancement methods for comparison include the non-learning based ones (NPE\cite{wang2013naturalness}, LIME\cite{guo2016lime}, RRM\cite{li2018structure}), ZeroDCE\cite{guo2020zero}, and the learning-based ones (RetinexNet\cite{Chen2018Retinex}, MBLLEN\cite{lv2018mbllen}, DSLR\cite{lim2020dslr}, KIND\cite{zhang2019kindling}, DRBN\cite{yang2020fidelity}, SNR-Aware\cite{xu2022snr}, LLFormer\cite{wang2023ultra} and MIRNet\cite{zamir2020learning}). The stereo-image low-light enhancement methods include DVENet\cite{huang2022low} and the newly proposed DCI-Net\cite{zheng2023decoupled}. Considering the limited number of the existing stereo-image low-light enhancement methods, we additionally introduced iPASSRNet\cite{wang2021symmetric} for comparison. Originally designed for stereo image super-resolution, iPASSRNet can be easily adapted as a stereo-image low-light enhancement model. With the two datasets, we re-trained iPASSRNet and evaluated its performance in our application.

\textbf{Implementation Details}. The hardware for the experiments is an Nvidia RTX 2080 Ti card with 12G memory. The batch size is set to 2, and the epoch number is set as 700 for training. We use the Adam optimizer to optimize with ${\beta _1}$=0.5 and ${\beta _2}$=0.999. The initial learning rate is set to 0.0001 and is reduced by half every 100 epochs.

\subsection{Quantitative Comparison}
The quantitative results of SDI-Net and the methods for comparison on Middlebury and Synthetic Holopix50k are reported in Table \ref{table1}. The non-learning traditional methods, such as NPE, LIME, JieP, and RRM, exhibit relatively poor performance. The reason is that it is difficult for the model-driven methods to fit the complex mapping functions between low-light and normal-light images. As for the learning-based monocular low-light enhancement methods, the PSNR and SSIM values are generally much higher, especially for the recently proposed methods such as DRBN and MIRNet. It is noted that the quantitative performance of ZeroDCE is relatively low. The reason is that it focuses on learning an optimal mapping curve function by only using dark images. The absence of full supervision information makes its performance not so well in terms of PSNR and SSIM. As for the stereo-image low-light enhancement, we can observe an obvious advantage over the monocular family, showing the effectiveness of using stereo images in general. Among these methods, our SDI-Net performs the best on both datasets in terms of PSNR and SSIM, showing the effectiveness and superiority of our method.

\subsection{Visual Comparison}
\textbf{Visual results on Middlebury}. Fig.\ref{fig4} and Fig.\ref{fig5} provide the visual comparisons for all the methods. In Fig.\ref{fig4}, we can see that some of the methods fail to improve the visibility of the input image, while other methods introduce color distortions or artifacts into their results. By comparing the zoomed-in patch of the enhanced results and the ground-truth image, we can see that our method performs better than others in terms of the visual quality, which well improves the visibility and recovers the detail textures simultaneously. We present another example in Fig.\ref{fig5}, in which we use error maps to visualize the discrepancies between the enhanced results and the ground truth \cite{Zheng2021WindowingDC}. From the error maps, we can clearly see that stereo-based methods performs better than the monocular family in general. Furthermore, our method is better than the other three stereo-based methods in handling fine details such as the brim of the straw hat region. The above observations is consistent with the trend of the above quantitative results.

\textbf{Visual results on Synthetic Holopix50k}. For the Synthetic Holopix50k dataset, we also utilize error maps to evaluate the enhanced results, as shown in Fig.\ref{fig6}. On one hand, the single-image low-light enhancement methods fail to effectively reconstruct normal-light images in terms of color and illumination. On the other hand, in comparison to other stereo vision methods, our approach outperforms them in reconstructing fine details, particularly in the leaves region. In addition, our method exhibits more continuous color textures and smoother image appearance in general.
\begin{table}[!ht]
\caption{Ablation studies of our model on the Middlebury dataset.}
    \centering
     \renewcommand{\arraystretch}{1.5}
    \resizebox{\linewidth}{!}{
    \normalsize
    \begin{tabular}{ccccc|c|c|c|c}
    \hline
        ~ & ~ & ~ & ~ & ~ & \multicolumn{2}{c|}{Left} & \multicolumn{2}{c}{Right} \\ \hline
        ~ & CAIM & PCAB & $\lambda$ & ${L_{fre}}$ & PSNR & SSIM & PSNR & SSIM \\ \hline
        V0 & × & × & 0.1 & $\surd$  & 31.47 & 0.9266 & 31.67 & 0.9212 \\ 
        V1 & × & $\surd$ & 0.1 & $\surd$ & 35.55 & 0.9405 & 35.71 & 0.9359 \\ 
        V2 & $\surd$ & × & 0.1 & $\surd$ & 34.49 & 0.9390 & 34.71 & 0.9349 \\ \hline
        V3 & $\surd$ & $\surd$ & × & × & 30.24 & 0.9006 & 30.04 & 0.8915 \\ \hline
        Ours & $\surd$ & $\surd$ & 0.1 & $\surd$ & 35.91 & 0.9501 & 35.99 & 0.9458 \\ \hline
    \end{tabular}
    }
\label{table2}
\end{table}
\subsection{Ablation study}
In the ablation study, we demonstrate the effectiveness of the CAIM module, PCAB module, and the employed loss functions, in which Middlebury is the evaluation dataset.

\textbf{Model structure}. In Table \ref{table2}, V0 represents the scenario where we perform a simple interaction on the extracted features in the middle of the encoders and decoders. In another word, the CSIM is replaced with a heuristic two-round interaction process. In the first round, $F_l$ and $F_r$ are directly concatenated and then down-sampled. In the second round, the down-sampled feature is respectively concatenated with $F_l$ and $F_r$ and down-sampled again. The refined feature representations are then sent into the decoding stage. V1 means that we add PCAB after the V0 version. V2 indicates that we only utilize CAIM for interaction but do not use the PCAB module. By comparing the performance of our complete model with V0 to V2, we observe that both CAIM and PCAB play indispensable roles in our low-light enhancement task. For example, the usefulness of CAIM are both empirically validated in the comparison between Ours and v1, and the comparison between V2 and V0. Similarly, the comparison between Ours and V2, and the comparison between V1 and V0 also empirically demonstrate the effectiveness of PCAB.

\textbf{Loss function}. In Table\ref{table2}, V3 means that the loss term ${L_{fre}}$ is not used. Compared to the completed model, we can see that there exists a clear performance decline for V3, which demonstrates the importance of using the loss term ${L_{fre}}$. The rationale is clear that the frequency-based image representation provides a complementary role to the spatial pixel arrays. Therefore, the combination of the two different loss terms effectively promotes the visual quality of enhanced results.

\section{Conclusion}
For the low-light image enhancement task, the research community have begun to explore the usage of more advanced imaging systems. For example, it is beneficial for enhancement models to simultaneously use the binocular images as the inputs. However, the limitation of insufficient interaction between these two views still exists in current research on low-light stereo image enhancement. In this paper, we propose SDI-Net for sufficiently modeling the dual-view interaction for low-light stereo image enhancement. By modeling the attention mechanism at the view level, the channel level and the pixel level, SDI-Net facilitates comprehensive information exchange of the dual views, therefore better restoring the high-quality normal-light images from low-light images. Experimental results on two public datasets demonstrate the effectiveness and superiority of SDI-Net in terms of quantitative and visual comparison. In the future, we plan to extend the interaction between the two views into the frequency domain and semantic domain.  

\bibliographystyle{IEEEtran}
\bibliography{IEEEabrv,SDINet}

\end{document}